\documentclass{llncs}
\usepackage[utf8]{inputenc}
\usepackage[T1,T2A]{fontenc}
\usepackage[hyphens]{url}
\usepackage[final]{graphicx}
\usepackage{epstopdf}

\begin{document}
	
\title{Morphological Analyzer and Generator for Russian and Ukrainian Languages}
\titlerunning{pymorphy2}  
%
\author{Mikhail Korobov}
%
\institute{ScrapingHub, Inc., Ekaterinburg, Russia\\ 
	\email{kmike84@gmail.com}}
\authorrunning{Mikhail Korobov} 
%
\tocauthor{Mikhail Korobov}

\maketitle              

\begin{abstract}
	pymorphy2 is a morphological analyzer and generator for Russian and Ukrainian languages. It uses large efficiently encoded lexicons built from OpenCorpora and LanguageTool data. A set of linguistically motivated rules is developed to enable morphological analysis and generation of out-of-vocabulary words observed in real-world documents. For Russian pymorphy2 provides state-of-the-arts morphological analysis quality. The analyzer is implemented in Python programming language with optional C++ extensions. Emphasis is put on ease of use, documentation and extensibility. The package is distributed under a permissive open-source license, encouraging its use in both academic and commercial setting.
	
\end{abstract}

\section{Introduction}
Morphological analysis is an analysis of internal structure of words. For languages with rich morphology like Russian or Ukrainian using the morphological analysis it is possible to figure out if a word can be a noun or a verb, or if it can be singular or plural. Morphological analysis is in an important step of natural language processing pipelines for such languages. 

Morphological generation is a process of building a word given its grammatical representation; this includes lemmatization, inflection and finding word lexemes.

pymorphy2 is a morphological analyzer and generator for Russian and Ukrainian language widely used in industry and in academia. It is being developed since 2012; Ukrainian support is a recent addition. The development of its predecessor, pymorphy\footnote{https://bitbucket.com/kmike/pymorphy} started in 2009. The package is available\footnote{https://github.com/kmike/pymorphy2} under a permissive license (MIT), and it uses open source permissively licensed dictionary data.

The rest of this paper is organized as follows. In Section 2 pymorphy2 software architecture and design principles are described. Section 3 explains how pymorphy2 uses lexicons and how analysis and morphological generation work for vocabulary words. In Section 4 methods used for out-of-vocabulary words are explained and compared with approaches used by other morphological analyzers. Section 5 is dedicated to a problem of selecting correct analysis from all possible analyses, and a role of morphological analyzer in this task. In Section 6 evaluation results are presented. Section 7 outlines a roadmap for future pymorphy2 improvements.

\section{Software Architecture}

pymorphy2 is implemented as a cross-platform Python\footnote{https://www.python.org/} library, with a command-line utility and optional C++ extensions for faster analysis. Both Python 2.x and Python 3.x are supported. An extensive testing suite (600+ unit tests) ensures the code quality; test coverage is kept above 90\%. There is online documentation\footnote{http://pymorphy2.readthedocs.org} available.

When optional C++ extension is used (or when pymorphy2 is executed using PyPy\footnote{http://pypy.org/} Python interpreter) the parsing speed is usually in tens of thousands of words per second; in some specific cases in can exceed 100000 words per second in a single thread. Without the extension parsing speed is in thousands of words per second.
The memory consumption is about 15MB, or about 30MB if we account for Python interpreter itself.

Users are provided with a simple API for working with words, their analyses and grammatical tags. There are methods to analyze words, inflect and lemmatize them, build word lexemes, make words agree with a number, methods for working with tags, grammemes and dictionaries. Inherent complexity of working with natural languages is not hidden from the user. For example, to lemmatize the word correctly it is necessary to choose the correct analysis from a list of possible analyses; pymorphy2 provides $P(analysis|word)$ estimates and sorts the results accordingly, but requires user to choose the analysis explicitly before normalizing the word.

Analysis of vocabulary words and out-of-vocabulary words is unified. There is a configurable pipeline of "analyzer units"; it contains a unit for vocabulary words analysis and units (rules) for out-of-vocabulary words handling. Individual units can be customized or turned off; some rules are parametrized with language-specific data. Users can create their own analyzer units (rules). This all makes it possible to perform morphological analysis experiments without changing pymorphy2 source code, develop domain-specific morphology analysis pipelines and adapt pymorphy2 to work with languages other than Russian. The latter point is validated by introducing an experimental support for Ukrainian language.

\section{Analysis of Vocabulary Words}

pymorphy2 relies on large lexicons for analysis of common words. For Russian it uses OpenCorpora \cite{bocharov13} dictionary ($\sim5*10^6$ word forms, $\sim0.39*10^6$ lemmas) converted from OpenCorpora XML\footnote{http://opencorpora.org/?page=export} format to a compact representation optimized for morphological analysis and generation tasks. End users don't have to compile the dictionaries themselves; pymorphy2 ships with prebuilt periodically updated dictionaries.

Any dictionary in OpenCorpora XML format can be used by pymorphy2. For Ukrainian there is such experimental dictionary ($\sim2.5*10^6$ word forms) being developed\footnote{Conversion utilities: https://github.com/dchaplinsky/LT2OpenCorpora} by Andriy Rysin, Dmitry Chaplinsky, Mariana Romanyshyn and other contributors; it is based on LanguageTool\footnote{https://languagetool.org/} data.

Source dictionary contains word forms with their tags, grouped by lexemes. For example, a lexeme for lemma "ёж" (a hedgehog) looks like this:

\begin{verbatim}
ёж      NOUN,anim,masc sing,nomn
ежа     NOUN,anim,masc sing,gent
ежу     NOUN,anim,masc sing,datv
...
ежами   NOUN,anim,masc plur,ablt
ежах    NOUN,anim,masc plur,loct
\end{verbatim}

In source dictionaries there could also be links between lexemes. For example, lexemes for infinitive, verb, gerund and participle forms of the same lemma may be connected. Currently pymorphy2 joins connected lexemes into a single lexeme for most link types.

\subsection{Morphological Analysis and Generation}

Given a dictionary, to analyze a word means to find all possible grammatical tags for a word. Obtaining of a normal form (lemmatization) is finding the first word form in the lexeme. To inflect a word is to find another word form in the same lexeme with the requested grammemes. 

As can be seen, all these tasks are simple. With an XML dictionary analysis of known words can be performed just by running queries on XML file. 

The problem is that querying XML is O(N) with large constant factors, raw data takes quite a lot of memory, and the source dictionary is not well suited for morphological analysis and generation of out-of-vocabulary words.

To create a compact representation and enable fast access pymorphy2 encodes lexeme information: all words are stored in a DAFSA \cite{daciuk00} using the dawgdic\footnote{https://code.google.com/p/dawgdic/} C++ library \cite{yata08} via Python wrapper\footnote{https://github.com/kmike/DAWG}; information about word tags and lexemes is encoded as numbers. Storage scheme is close to the scheme described in aot.ru \cite{sokirko04}, but it is not quite the same.

\subsubsection{Paradigms}

Paradigm in pymorphy2 is an inflection pattern of a lexeme. It consists of $prefix_i, suffix_i, tag_i$ triples, one for each word form in a lexeme, such as that each word form $i$ can be represented as $prefix_i + stem + suffix_i$ where $stem$ is the same for all words in a lexeme. 

This representation allows us to factorize a lexeme into a stem and a paradigm. 

Paradigm prefixes, suffixes and tags are encoded as numbers by pymorphy2; lexeme stems are discarded. It means that a paradigm is stored as an array of numbers (prefixes, suffixes and tags IDs), and lexemes are not stored explicitly - they are reconstructed on demand from word and paradigm information.

There are no paradigms provided in the source dictionary; pymorphy2 infers them from the lexemes. For Russian there are about 3200 paradigms inferred from 390000 lexemes.

\subsubsection{Word Storage}

Word forms with their analysis information are stored in a DAFSA. Other storage schemes were tried, including two tries scheme similar to described in \cite{segalovich03} (but using double-array tries), and succinct (MARISA\footnote{https://code.google.com/p/marisa-trie/}) tries. For pymorphy2 data DAFSA provided the most compact representation, and at the same time it was the fastest and had the most flexible iteration support.

\begin{figure}[h]
    \centering
    \includegraphics[width=12cm]{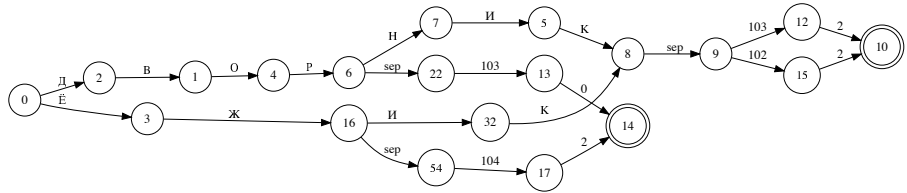}
    \caption{DAFSA encoding example. Encoded (word, paradigmId, formIndex) triples: (двор, 103, 0);  (ёж, 104, 0); 
        (дворник, 101, 2); (дворник, 102, 2); 
        (ёжик, 101, 2); (ёжик, 102, 2)}
    \label{fig:DAFSA}
\end{figure}

For each word form pymorphy2 stores (word, paradigmId, formIndex) triples:

\begin{itemize}
    \item word form, as text;
    \item ID of its paradigm;
    \item word form index in the lexeme.
\end{itemize}

DAFSA doesn't support attaching values to leaves; the information is encoded like the following: $<word>SEP<paradigmId><formIndex>$ (see an example on fig. \ref{fig:DAFSA})\footnote{pymorphy2 encodes words to UTF-8 before putting them to DAFSA, so in practice there are more nodes than shown on fig. \ref{fig:DAFSA}. It is an implementation detail.}.

The storage is especially efficient because words with similar endings often have the same analyses, i.e. the same $(paradigmId, formIndex)$ pairs; this allows DAFSA to use fewer nodes/transitions to represent the data. DAFSA for Russian OpenCorpora dictionary ($5*10^6$ analyses, about $3*10^6$ unique word forms) enables fast lookups (hundreds thousand lookups/sec from Python) and takes less than 7MB of RAM; source XML file is about 400MB on disk.

To get all analyses of a $word$, DAFSA transitions for $word$ are followed, then a separator $SEP$ is followed, and then the remaining subtree is traversed to get all possible $(paradigmId, formIndex)$ pairs. 

Given $(paradigmId, formIndex)$ pair one can find the grammatical tag of a word: find a paradigm in paradigms array by $paradigmId$, get $(prefix_i, suffix_i, tag_i)$ triple from a paradigm by using $i:=formIndex$. Given $(paradigmId, formIndex)$ pair and the $word$ itself it is possible to restore the lexeme and lemmatize or inflect the $word$ - from $word$, $prefix_i$ and $suffix_i$ we can get the stem, and given a stem and $(prefix_k, suffix_k, tag_k)$ it is possible to restore a full word for k-th word form.

\subsection{Working with "ё" and "ґ" Characters Efficiently}

The usage of "ё" letter is optional in Russian; in real texts it is often replaced with "е" letter. There rules for "ґ" / "г" substitutions are different in Ukrainian, but in practice there are real-world texts with "ґ" letters replaced with "г". 

The simplest way to handle it is to replace "ё" / "ґ" with "е" / "г" both in the input text and in the dictionary. However, this is suboptimal because it discards useful information, makes the text less correct (in Ukrainian "г" instead of "ґ" can be seen as a spelling error) and increases the ambiguity: there are words which analysis should depend on е/ё and ґ/г. For example, the word "все" should be parsed as plural, but the word "всё" shouldn't.

pymorphy2 assumes that "ё" / "ґ" usage in dictionary is mandatory, but in the input text it is optional. For example, if a Russian input word contains "ё" letter then only analyses with this letter are returned; if there are "е" letters in the input word then possible analyses both for "е" and "ё" are returned.

An easy way to implement this would be to check each combination of е/ё and ґ/г replacement for the input word. It is not how pymorphy2 works. To do the task efficiently, pymorphy2 exploits DAFSA \cite{daciuk00} dictionary structure: the result is built by traversing the word character graph and trying to follow "ё" transitions in addition to "е" transitions (for Russian) and "ґ" transitions in addition to "г" transitions (for Ukrainian).

\section{Analysis of Out-of-Vocabulary Words}

It is not practical to try incorporate all the words in a lexicon - there is a long tail of rarely used words, new words appear; there is morphological derivation, loanwords, it is challenging to add all names, locations and special terms to the dictionary. Empirically, Zipf's Law seems to hold for natural languages \cite{zipf32}; one of the consequences is that even doubling the size of a lexicon could increase the coverage only slightly \cite{daciuk99}.

For languages without rich morphology it may be practical to assume that if word is not in a dictionary then it can be of any class from the open word classes, and then disambiguate the results on later processing stages, using e.g. a contextual POS tagger or a syntactic parser. For Slavic languages doing this on later stages is challenging because of large tagsets - for example, OpenCorpora \cite{bocharov13} words have more than 4500 different tags. Morphological analyzer solves it by limiting the number of possible analyses based on word shape.

pymorphy2 uses a set of rules (analyzer units) to handle unknown words. Some of the rules are described in literature \cite{mikheev97,segalovich03,sokirko04,daciuk99,bolshakov12}; the resulting combination is novel. The order of in which the rules are applied is language-specific.

\subsection{Common Prefixes Removal}

There is a set of immutable prefixes which can be attached to words of open classes (nouns, verbs, adjectives, adverbs, participles, gerunds) without affecting the word grammatical properties. Examples of such prefixes for Russian: "не", "псевдо", "авиа"; pymorphy2 provides language-specific lists of these prefixes. 

When a words starts with one of these prefixes, pymorphy2 removes the prefix, parses the reminder and re-attaches the prefix. A similar rule is described in \cite{mikheev97}. Note that full analysis is performed on the reminder, so the reminder can be an out-of-vocabulary word itself. To speedup prefix matching built-in lists of prefixes are encoded to DAFSAs.

\subsection{Words Ending with Other Dictionary Words}

When all the following apply pymorphy2 assumes the whole word can be parsed the same way as the "suffix" word:

\begin{itemize}
    \item a word being analyzed has another word from a dictionary as a suffix;
    \item the length of this "suffix" word is greater than 3;
    \item the length of the word without the "suffix" is no greater than 5;
    \item "suffix" word is of an open class (noun, verb, adjective, participle, gerund)
\end{itemize}

To search for suffixes pymorphy2 tries to consider 1st letter as a prefix, then two first letters as a prefix, etc., and lookups the reminder in a dictionary.

This rule is the same as described in \cite{sokirko04}. A similar rule is described in \cite{mikheev97}, though its induction for concrete prefixes is different. 

\subsection{Endings Matching}

In many languages, including Russian and Ukrainian, words with common endings often have the same grammatical form. 

To exploit this, pymorphy2 first collects the information from the dictionary: for each word all endings of length 1 to 5 are extracted, and all possible analyses for these endings are stored. Then this $ending \rightarrow \{analyses\}$ mapping is cleaned up:

\begin{itemize}
    \item only the most frequent analyses for each POS tag are kept;
    \item analyses from non-productive paradigms (currently these are paradigms which produced less than 3 lexemes in a dictionary) are discarded;
    \item rare endings (currently the ones which occur once) are also discarded. 
\end{itemize}

The resulting mapping is encoded to DAFSA for fast lookups. Storage scheme is the following: $<ending>SEP<analysisInfo>$, where $analysisInfo$ consists of three 2-byte numbers: $(frequency, paradigmId, formIndex)$ - analysis frequency (a number of times a word with this ending had this analysis), ID of analysis paradigm and the form index inside the paradigm.

At prediction time pymorphy2 checks word endings from length 5 to 1, stopping at the first ending with some analyses found. To get possible analyses for a given ending pymorphy2 first follows all DAFSA transitions for the ending, then follows a separator, and then traverses the remaining subtree to get possible $analysisInfo$ triples. The result is then sorted by analyses frequencies. 

Recall that a $word$ and a $(paradigmId, formIndex)$ pair is all what is needed to restore the lexeme and inflect the $word$. Lexemes are created on fly, so it doesn't matter $word$ is not from the vocabulary as soon as we have $(paradigmId, formIndex)$ pair. It means morphological generation (lemmatization, inflection) works here.

Only analyses with open-class parts of speech (noun, verb, adjective, participle, gerund) are produced. Special care is taken to handle "ё" letter properly. Also, special care is required to handle paradigm prefixes properly - in fact, there are several $ending \rightarrow \{analyses\}$ DAFSAs built, one per each paradigm prefix.

This rule is based on \cite{sokirko04}; similar approaches are also used in \cite{bolshakov12} and \cite{segalovich03}. \cite{mikheev97} uses similar rules, but derives them differently.

\subsection{Words with a Hyphen}

Unlike some other morphological analyzers, pymorphy2 opts to handle words with a hyphen.

In \cite{krylov03} it is argued that in most cases the parts of compound words should be handled as separate words if they are joined using a hyphen. A similar decision is made in OpenCorpora tokenization module \cite{bocharov11}; it considers words like "Жан-Поль" as three tokens which should be analyzed separately and joined back at later processing stages. In both cases the decisions are not motivated by linguistic considerations; it is the technical difficulty which prevents analyzing and processing such words as single entities.

Currently pymorphy2 handles adverbs with a hyphen, particles separated by a hyphen and compound words with left and right parts separated by a hyphen.

\subsubsection{Adverbs with a Hyphen}

Russian words are parsed as adverbs if they

\begin{itemize}
    \item start with a "по-" prefix;
    \item have total length greater than 5;
    \item can be parsed as a full singular adjective in dative case when "по-" is removed 
\end{itemize}

 Examples: "по-северному", "по-хорошему".

\subsubsection{Particles Separated by a Hyphen}

Though it is not clear if words with a particle separated by a hyphen (e.g. "смотри-ка" or "посмотрел-таки") should be handled as a single word or as two words, pymorphy2 supports parsing of such words. There are language-specific lists of common particles which can be attached, and if a word ends with one of these particles then it is parsed without the particle, and then the particle is re-attached to the result.

\subsubsection{Compound Words with a Hyphen}

The main challenge in analysis of the compound words which parts are separated by a hyphen (like "человек-паук" and "Царь-пушка") is to figure out if the left part should be inflected together with the right part, or if it is a fixed prefix.

To do this, pymorphy2 parses left and right parts separately (they don't have to be dictionary words). Then it tries to find matching analyses. If there is a "left" analysis compatible with one of the "right" analyses then the resulting analysis is built where both word parts are inflected. After that, an analysis with a fixed left part is added to the result, regardless of whether a compatible "left" analysis was found or not. A similar method was used in \cite{bolshakov12}.

Only words with a single hyphen are handled using heuristics described above. Words with multiple hyphens are likely represent different phenomena in Russian and Ukrainian languages; they could be interjections or phrases \cite{zanegina12}.

\subsection{Other Tokens}

Initial is an abbreviation of person's first or patronymic name. In most cases an initial is a single upper-cased character (language-specific). pymorphy2 parses such characters as fixed singular nouns, with variants for all possible gender and case combinations. For person first names ($Name$) two different lexemes are built for male and female names. For patronymic names ($Patr$) a single lexeme is returned. Unlike all other analyzer rules, detection of initials is case-sensitive. It is a way to decrease ambiguity. 

The following tags are assigned to non-lexical tokens: $PNCT$ for punctuation, $LATN$ for tokens written in Latin alphabet, $NUMB,intg$ for integer numbers, $NUMB,real$ for floating-point numbers, $ROMN$ for Roman numbers.

When analyzing the text, it is common to classify tokens during the tokenization step. The reason pymorphy2 handles non-lexical tokens during the morphological analysis step is that this allows users to use a simpler tokenizer when classifying tokenizer is not available; also, it means that information about all tokens is available in a common format. 

\subsection{Morphological Generation of Out of Vocabulary Words}

Inflection is fully supported for out of vocabulary words. To achieve this pymorphy2 keeps track of the analyzer units (rules and their parameters) used to parse the word, requires each analyzer unit to provide a method for getting a lexeme, and calls this method for the last analyzer unit. To compute the lexeme analyzer unit can look at the analysis result, and it can ask previous analyzer units for the lexeme.

For example, Common Prefixes Removal analyzer removes the prefix from a word, then gets a lexeme from the previous analyzer, and then attaches the prefix to each word form in a lexeme to build a resulting lexeme.

\section{Probability Estimation}

Morphological analyzer may return multiple possible word parses. The problem of choosing the right analysis from a list of possible options is called disambiguation. Generally, to select the correct analysis it is required to take word context in account. Morphological analyzer takes individual words as an input, so it can't disambiguate the result robustly. However, it can provide an estimation for $P(analysis | word)$ conditional probability. Such probability estimations can be used in absence of a dedicated disambiguator to select the more probable analysis. In addition to that, these probabilities can be used on later stages of text analysis, for example by a disambiguator.

To estimate $P(analysis | word)$ conditional probability for Russian words pymorphy2 uses  partially disambiguated OpenCorpora corpus \cite{bocharov13}  and assumes that $P(analysis | word) = P(tag | word)$. The conditional probability is estimated for words which have multiple analysis according to pymorphy2, but have occurrences with a single remaining analysis in the OpenCorpora corpus; the estimation is a maximum-likelihood estimation with Laplace (add-one) smoothing.

$$W_{disambiguated} := \{word: |tags_{corpus}(word)| = 1, word \in corpus \}$$
$$W_{ambiguous} :=   \{word: |tags_{pymorphy2}(word)| > 1, word \in W_{disambiguated}\} $$
$$B(word) = \max(|tags_{pymorphy2}(word)|, |tags_{corpus}(word)|)$$

$$\forall word \in  W_{ambiguous},$$
$$\forall tag \in tags_{pymorphy2}(word):$$
\begin{equation}
P_{MLE}(tag | word) = \frac
	{count(word, tag) + 1}
	{count(word) + B(word)}
\end{equation}

Counts are computed based on OpenCorpora corpus data; all words with a single remaining analysis are taken in account.

Once estimated, the result is stored on disk as a DAFSA; keys are
$$<word>:<tag><NULL><int(10^6 * P_{MLE}(tag | word))>$$

For words without $P_{MLE}(tag | word)$ estimates the probabilities are assigned uniformly during the parsing.

For Ukrainian language probabilities are assigned uniformly because at the moment of writing there is no a freely available Ukrainian corpus similar to OpenCorpora.

\section{Evaluation}

Evaluating analysis quality of different morphological analyzers for Russian is not straightforward because most analyzers (as well as annotated corpora) use their own incompatible tagsets. And when a corpus and a dictionary have a compatible tagset it usually means that the dictionary was enhanced from the corpus, and it is a problem because quality numbers obtained on a corpus the dictionary was enhanced from shouldn't be relied on - they are too optimistic.

pymorphy2 analysis quality was compared to an analysis quality of a well-known morphological analyzer\footnote{https://tech.yandex.ru/mystem/}, Mystem 3.0 \cite{segalovich03}. Testing corpus consists of 100 randomly selected sentences (1405 tokens) from OpenCorpora (microcorpus\footnote{https://github.com/kmike/microcorpus}) and 100 randomly selected sentences (1093 tokens) from ruscorpora.ru - 2498 manually disambiguated tokens in total.

Full details for this evaluation can be found online\footnote{http://nbviewer.ipython.org/gist/kmike/52fb0a9b3ed627310bea}. 

OpenCorpora (pymorphy2) tagset is not the same as ruscorpora.ru tagset, and ruscorpora.ru tagset differs from Mystem tagset. For evaluation purposes all tags were converted to Mystem format using a set of automatic rules. Quality was evaluated on full morphological tags, i.e. tags must match exactly to be considered correct, with a few exceptions related to tags conversion problems.
All reported errors were checked manually to filter out false positives.

\begin{table}[h]
    \caption{Errors}
    \centering
    \label{tab:representation}    
    \begin{tabular}{|r|l|l|}
        \hline
        & pymorphy2 & mystem 3.0 \\ \hline
        microcorpus  & 10        & 15         \\ \hline
        ruscorpora   & 9         & 8          \\ \hline
        total               & 19        & 23         \\ \hline
    \end{tabular}
\end{table}

Both pymorphy2 and Mystem made less than 1\% errors (without disambiguation, i.e. in less than 1\% cases the correct analysis was not in a set of analyses returned by an analyzer). It should be noted that 9 out of 19 pymorphy2 errors and 14 out of 23 Mystem errors were related to abbreviation handling. Mystem handled first and last names better (1 mistake versus 4 for pymorphy2); pymorphy2 made less mistakes for "regular" words (4 versus 6 for mystem). Mystem can't parse many hyphenated words as a single token; such words were not considered. Punctuation, numbers and non-Russian words were also removed from the input.

It is hard to draw a quantitative conclusion because the corpus size is small. Both analyzers has a similar analysis quality, and the resulting numbers depend on evaluation minutiae: whether abbreviations are considered or not, should we require hyphenated words to be parsed, do we require verb transitivity to be predicted correctly, is it important to distinguish adverbs from parenthesis, etc.

Several human annotation errors were found by parsing OpenCorpora data with mystem (1 error) and ruscorpora data with pymorphy2 (6 errors). OpenCorpora shares a dictionary with pymorphy2, and ruscorpora annotation is related to mystem; this shows an utility of using cross-corpora tools to check the annotations.

The most sophisticated Russian morphological parser evaluation so far is \cite{rueval10}; it happened in 2010. Previous version of pymorphy2 (pymorphy) participated\footnote{Anonymized results: http://ru-eval.ru/tables\_index.html} in tracks without disambiguation; it finished 1st on Full Morphology Analysis, 3rd on Lemmatization, 3rd on POS tagging and 5th on the Rare Words track. pymorphy haven't participated in disambiguation tracks.

pymorphy used some pymorphy2 rules (not all) and a different dictionary (extracted from \cite{sokirko04} instead of \cite{bocharov13}). Generally, pymorphy2 should work better than pymorphy because of an improved dictionary and rules, but this has not been not measured quantitatively yet.

\section{Conclusion and Future Plans}

Permissive open-source license (MIT) is used for pymorphy2. All the dictionaries and corpora pymorphy2 depends on are also available under permissive open-source licenses. This encourages usage and contributions. There are volunteers working on Russian and Ukrainian dictionaries and corpora, related tools and pymorphy2 itself.

Development of pymorphy2 is by no means finished. There are word classes for which pymorphy2 analysis can be improved. Some of them: people last and patronymic names, foreign people names, diminutive first names, locations, uppercase and other abbreviations, some classes of hyphenated words, ordinal numbers (including ordinal numbers written in digit notation like "22-й"). According to \cite{rueval10}, similar issues are common for Russian morphological analyzers.
  
 Non-contextual $P(tag | word)$  estimates can be made better by transferring some information about similar words and by improving the corpora.
 
A better comparison between pymorphy, pymorphy2, Mystem and other morphological analyzers could require a robust tagset conversion library.
  
 The support for Ukrainian is experimental. The dictionary requires work, pymorphy2 needs more Ukrainian-specific rules for handling of out of vocabulary words, and for better $P(tag | word)$  estimates an annotated Ukrainian corpus is needed: even a small corpus (or even a manually crafted frequency list) should fix a substantial amount of "obvious" errors.
 
 There are plans to add Belarusian language support to pymorphy2 based on Belarusian N-korpus\footnote{http://bnkorpus.info} grammar database.
 
Although pymorphy2 is already fast enough for many use cases (tens of thousands words per second in a single thread), there is a room for further speed improvements.

%
%
\newpage
\bibliographystyle{splncs}

\end{document}